%% file: main.tex
\documentclass[titlepage,oneside,letterpage,12pt]{article}

\usepackage{lscape}
\usepackage{algorithm,algorithmic}

\input{pream.tex}

\newcommand{\xx}{{the congested network property }}
\newcommand{\ta}{{\sc REINFORCE-TD}}

\begin{document}
\setcitestyle{square} 

\thispagestyle{empty}
\begin{titlepage}
\begin{flushleft}
{\MakeUppercase{\bfseries 
 Large-scale traffic signal control using machine learning:  some traffic flow considerations
}}\\[36pt]

{\bfseries Jorge A. Laval, Corresponding Author}\\
Georgia Institute of Technology\\
790 Atlantic Drive, Atlanta, GA 30332\\
Tel: 404-894-2360 Email: jorge.laval@ce.gatech.edu\\[60pt]

{\bfseries Hao Zhou}\\
Georgia Institute of Technology \\
790 Atlantic Drive, Atlanta, GA 30332\\
Tel: tel Email: email\\[12pt]

Word Count: 4989 words + 10 fig(s) = 7489~words\\[12pt]

\end{flushleft}
\end{titlepage}


\newpage
\section{Abstract}
This paper uses supervised learning, random search and deep reinforcement learning (DRL) methods to control large signalized intersection networks. The control policy at each intersection is 
parameterized as a deep neural network, to approximate the ``best'' signal setting  as a function of the state of its incoming and outgoing approaches. 
The traffic model is Cellular Automaton rule 184, which has been shown to be a parameter-free representation of traffic flow, and which is probably the most efficient implementation of the Kinematic Wave model with triangular fundamental diagram.  We are interested  in the steady-state performance of the system, both spatially  and temporally: we consider a homogeneous grid network inscribed on a torus, which makes the network boundary-free,  and drivers choose random routes.  
As a benchmark we use the longest-queue-first (LQF) greedy \citep{arel2010reinforcement} algorithm.
We find that:
\begin{enumerate*}[label=(\roman*)]
    \item a policy trained with supervised learning with only two examples outperforms LQF,  
    \item random search is able to generate near-optimal policies,
      \item the prevailing average network occupancy during training is the major determinant of the effectiveness of  DRL policies. When trained  under free-flow conditions one obtains DRL policies that are optimal for all traffic conditions, but this performance deteriorates as the occupancy during training increases. For occupancies $> 75\%$ during training, DRL policies perform very poorly for all traffic conditions, which means that DRL methods cannot learn under highly congested conditions.
\end{enumerate*}

We conjecture that DRL's inability to learn under congestion might be explained by
a property of urban networks found here, whereby even a very bad policy produces an intersection throughput higher than downstream capacity. This means that the actual throughput tends to be independent of the policy. 
Our findings imply that it is advisable for
  current DRL methods in the literature to discard any congested data when training, and that doing this will improve their performance under all traffic conditions. 
  They also suggest that this inability to learn under congestion might be
  alleviated by combining DRL for free-flow and supervised learning for congested conditions.
\\
\\
\\
\\
\textit{Keywords}: Traffic signal control, machine learning, deep reinforcement learning
\newpage

\section{Introduction}

The use of deep neural networks  within Reinforcement Learning  algorithms has produced important breakthroughs in recent years. These deep reinforcement learning (DRL) methods have outperformed expert knowledge methods in areas such as arcade games, backgammon, the game  Go  and autonomous driving \citep{mnih2015human, silver2017mastering,chen2019model}. 
In the area of traffic signal control  numerous DRL control methods have been proposed both for isolated intersections \cite{li2016traffic,genders2016using} and small networks \citep{chu2015traffic,chu2019multi,tan2019cooperative,ge2019cooperative}.  The vast majority of these methods have been trained with a single (dynamic) traffic demand profile, and then validated using another one, possibly including a surge \citep{ge2019cooperative}. 

A gap in the literature appears to be a consistent analysis of the different aspects of large traffic flow networks that influence the performance of DRL methods. For example, it is not clear if and how network congestion levels affect the learning process, or if other machine learning methods are effective, or if current findings also apply to large networks. This paper is a step in this direction, where we  examine the simplest  possible DRL  setup in order to  gain some insight  on how the optimal policy changes  with respect to different configurations of the learning framework.  In particular, we are interested  in the steady-state performance of the system, both spatially  and temporally: we consider a homogeneous grid network inscribed on a torus, which makes the network boundary-free,  and drivers choose random routes.  

In the current signal control DRL literature the problem is treated, invariably, as an episode process, which is puzzling given that the problem is naturally a continuing (infinite horizon) one. Here, we adopt the \textit{continuing} approach to maximize the long-term average reward. We argue that in signal control there is no terminal state because the process actually goes on forever. And what may appear as a terminal state, such as an empty network, cannot be considered so because it is not achieved through the correct choice of actions but by the traffic demand, which is uncontrollable. An explanation for this puzzling choice in the literature might be that DRL training methods for episodic problems have a much longer history and our implemented in most machine learning development frameworks.
For continuing problems this is not unfortunately the case, and we propose here the training algorithm \ta, which is in the spirit of REINFORCE  with baseline \citep{willianms1988toward} but for continuing problems. To the best of our knowledge, this extension of REINFORCE is not available in the literature. 

The remainder of the paper is organized as follows. We start with the background section regarding DRL methods and the macroscopic fundamental diagram of urban networks, followed by a survey of related work. Then, we define the problem set up and apply it to a series of experiments that highlight the main properties found here. Finally, the paper concludes with a discussion and outlook section.

\section{BACKGROUND}

\subsection{The macroscopic fundamental diagram (MFD) of urban networks}

Macroscopic models for traffic flow have become increasingly popular after 
the empirical verification of a network-level Macroscopic Fundamental Diagram (MFD) on congested urban areas \citep{Daganzo2007Urban,Geroliminis2008Existence}. 
For a given traffic network, the  MFD describes the relationship between traffic variables averaged across all \textit{lanes} in the network. In this paper we will use the flow-density MFD, which gives the average flow on the network as a function of the average density on the network.

The main requirement for a well-defined MFD is that congestion be homogeneously distributed across the network, i.e. there must be no ''hot spots'' in the network. For analytical derivations it is often also assumed that each \textit{lane} of the network obeys the  kinematic wave model \citep{Lighthill1955Kinematic,richards1956shock} with common fundamental diagram \citep{daganzo2008analytical, Laval2015Stochastic}. 
In this way, upper bounds for the MFD have been found using the method of cuts in the case of homogenous networks. For general networks,  \cite{Laval2015Stochastic} show that (the probability distribution of) the MFD can be well approximated by a function of mainly two  parameters: the mean distance between traffic lights divided by the mean green time, and the mean red-to-green ratio across the network.

\subsection{Reinforcement learning}

Reinforcement learning is typically formulated within the framework of a {\em Markov
decision process} (MDP). At discrete time step $t$ the environment is in state $S_t\in{\cal S}$, the agent will choose and action  $A_t\in{\cal A}$, to maximize a function of future rewards $R_{t+1}, R_{t+2}\ldots$ with $R_-: {\cal S} \times {\cal A} \rightarrow \Re$. There is a state transition probability distribution $P(s',r|s,a)=\Pr(S_t=s',R_t=r|S_{t-1}=s, A_{t-1}=a)$ that gives the probability of making a transition from state $s$
to state $s'$ using action $a$ is denoted $P(s,a,s')$, and is commonly referred to as the ``model''.
The model is
{\em Markovian} since  the state transitions are independent of any previous
environment states or agent actions. For more details on MDP  models the reader is referred to \cite{bellman1957markovian,bertsekas1987dynamic,howard1960dynamic,puterman1994markovian}

The agent's decisions are characterized by a stochastic
\textbf{policy} \( \pi (a|s) \), which is the probability of taking action
\( a \) in state \( s \).
In the continuing case the agent seeks to maximize the \textit{average reward}:
\begin{equation}\label{eta}
    \eta (\pi )\equiv \lim_{T\rightarrow\infty} \frac{1}{T}\sum_{t=1}^T E_{\pi}\left[R_t\right]
\end{equation}
 The term $E_{\pi}$ means that the expected value (with respect to the distribution of states) assumes that the policy is followed.

In the case of traffic signal control for large-scale grid network, methods based on transition probabilities are impractical because the state-action space tends to be too large as the number of agents increases.
An alternative approach that circumvents this  \textit{curse of dimensionality} problem---the approach we pursue here---are ``policy-gradient'' algorithms,  where the policy is parameterized as \( \pi (a|s;{\theta }), \theta \in \mathcal{R}^{m} \), typically a neural network. Parameters $\theta$ are adjusted to 
improve the performance of the policy $\pi$ by following the gradient of cumulative
future rewards, given by the identity
\begin{equation}
\label{policy_grad}
\nabla \eta=E_{\pi}[G_t \nabla_{\theta}\log  \pi (a|s)]
\end{equation}
as shown in \cite{sutton1999policy} for both continuing and episodic problems. In continuing problems cumulative rewards $G_t$ are measured relative to the average cumulative reward:
\begin{equation}\label{return}
    G_t=\sum_{i=t+1}^\infty (R_i-\eta(\pi))
\end{equation}
and is known as  the \textit{differential return}.
The value function is the
expected differential return  the agent will gain if
it starts in that state and executes the  policy  $\pi$.
\begin{equation}
V(s) = E_{\pi}[G_t |S_t=s]
\label{e:vstar}
\end{equation}  

\subsection{Related work}


The existing literature is split between two approaches for formulating the large-scale traffic control problem: 
either a centralized DRL algorithm  or a decentralized method with communication and cooperation among multi-agents. 
The centralized approach \citep{genders2016using,li2016traffic,chu2016large} usually adopts a single-agent learning algorithm as many DRL control problems and tries to tackle the high-diamentional continuous control problem by novel algorithms like memory replay, dual networks and advantage actor-critic \citep{lillicrap2015continuous,mnih2015human}. The decentralized method takes advantage of multiple agents and requires design of efficient communication and coordination to address the limitation of partial observation of local agents. Current studies \citep{khamis2014adaptive,wei2019colight,tan2019cooperative,gong2019decentralized} often decompose the large network into small regions or individual intersections, and train the local-optimum policies separately given reward functions reflecting certain level of cooperation. It is worth noting that different observational measures of the environment are used as communication information between agents, such as neighbouring intersections, downstream intersections or upstream intersections. How to incorporate those communication information to help design the reward function for local agents remains an open question.






The environment modeling, state representation and reward function design are key ingredients in DRL. For the environment emulator, most studies are based on popular microscopic traffic simulation packages like AIMSUM or SUMO. Recently, FLOW \cite{kheterpal2018flow} has been developed as a computational framework integrating SUMO with some advanced DRL libraries to implement DRL algorithm on ground traffic scenarios. \citep{vinitsky2018benchmarks} provided a benchmark for major traffic control problems including the multiple intersection signal timing. 
There also exist studies \cite{chu2015traffic,arel2010reinforcement,ge2019cooperative} adopting methods to use self-defined traffic models as the environment. Complementary to those microscopic simulation packages, macroscopic models are able to represent the traffic state using cell or link flows. The advantage of macroscopic models is twofold: i) reducing complexity in state space and computation ii) being compatible with domain knowledge from traffic flow theory such as MFD theory.

Expert knowledge has been included in some studies  to reduce the scale of the network control problem. In \cite{xu2018network}, critical nodes dictating the traffic network were identified first before the DRL was implemented. The state space can be remarkably reduced. Macroscopic fundamental diagram (MFD) theory cannot provide sufficient information to determine the traffic state of a network. For instance, \cite{chu2015traffic} successfully integrated the MFD with a microscopic simulator to constrain the searching space of the control policies in their signal design problem. They defined the reward as the trip completion rate of the network, and simultaneously enforcing the network to remain under or near the critical density. The numerical experiments demonstrated that their policy trained by the integration of MFD yields a more robust shape of the MFD, as well as a better performance of trip completion maximization, compared to that of a fixed and a greedy policy.



While most of the related studies on traffic control only focus on developing effective and robust deep learning algorithms, few of them have shown traffic considerations, such as the impact of traffic density. The learning performance of RL-based methods under different densities have not been sufficiently addressed. To the best of our knowledge, \cite{camponogara2003distributed} is the only study which trained a RL policy for specific and varied density levels, but unfortunately their study only accounted for free-flow and mid-level congestion. 
\cite{dai2011neural} classified the traffic demand into four vague levels and reported that inflow rates at 1000 and 1200 veh/h needed more time for the algorithm to show convergence. But they did not report network density, nor try more congested situations nor discussed why the converging process has been delayed. 
Most studies only trained RL methods in non-congestion conditions, \cite{ge2019cooperative} adopted the Q-value transfer algorithm (QTCDQN) for the cooperative signal control between a simple 2*2 grid network and validated the adaptability of their algorithm to dynamic traffic environments with different densities, such as the the recurring congestion and occasional congestion. 


In summary, most recent studies focus on developing effective and robust multi-agent DRL algorithms to achieve coordination among intersections. The number of intersections in those studies are usually limited, thus their results might not apply to large open network. Although the signal control is indeed a continuing problem, it has been always modeled as an episodic process. From the perspective of traffic considerations, expert knowledge has only been incorporated in down-scaling the size of the control problem or designing novel reward functions for DRL algorithm. Few studies have tested their methods given different traffic demands, or shed lights on the learning performance under different traffic conditions, especially the congestion regimes. To fill the gap, our study will treat the large-scale traffic control as a continuing problem and extend classical RL algorithm to fit it. More importantly, noticing the lack of traffic considerations on learning performance, we will train DRL policies under different density levels and explore the results from a traffic flow perspective. 




\section{Problem set up}


\textbf{The traffic flow model} used in this paper is the kinematic wave model \citep{Lighthill1955Kinematic,richards1956shock}  with a triangular flow-density fundamental diagram, which is the simplest model able to predict the main features of traffic flow. The shape of the triangular fundamental diagram  is irrelevant  due to a symmetry in the kinematic wave model  whereby flows and delays are invariant  with respect to linear transformations,  and renders the kinematic wave model parameter-free; see \cite{Lav16} for the details.  This allows us to use  an isosceles fundamental diagram, which in combination with a cellular automaton (CA) implementation  of the kinematic wave model, produces its most computationally efficient  numerical solution method: Elementary CA Rule 184 \cite{wolfram1984cellular}.  

\begin{figure}[tb]
\centering
\includegraphics[width=.6\textwidth]{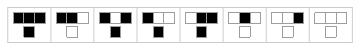}
\caption{CA Rule 184: The top row in each of the eight cases  shows the  the neighborhood values $( c_{i-1} ,  c_i , c_{i+1})$  and the updated $c_i$ in the bottom row.
}
\label{f1}
\end{figure}

In a CA model,  each lane of the road is divided into  small cells  $i=1,2,\ldots n$ the size of a vehicle jam spacing, where cell $n$ is the most downstream  cell of the lane. The value in each cell, namely $c_i$, can be either ``1''  if a vehicle is present and ``0'' otherwise. The update  scheme  for CA Rule 184, shown in Fig.~\ref{f1}, operates over a neighborhood   of length 3, and can be written as:
\begin{equation}\label{CA Rule 184}
c_i :=c_{i-1}\lor c_{i-1}\land c_i \lor c_i\land c_{i+1} 
\end{equation}
The vector $c$ is a vector of bits  and \eqref{CA Rule 184} is Boolean algebra, which explains the high computational efficiency of this traffic model. Notice that \eqref{CA Rule 184} implies that the current state of the system is described completely by the state in the previous time step; i.e. it is Markovian and deterministic. Stochastic components are added by the signal control policy, and therefore our traffic model satisfies the main assumption of the MDP framework. 
 
\begin{figure}[tb]
\centering
\includegraphics[width=.7\textwidth]{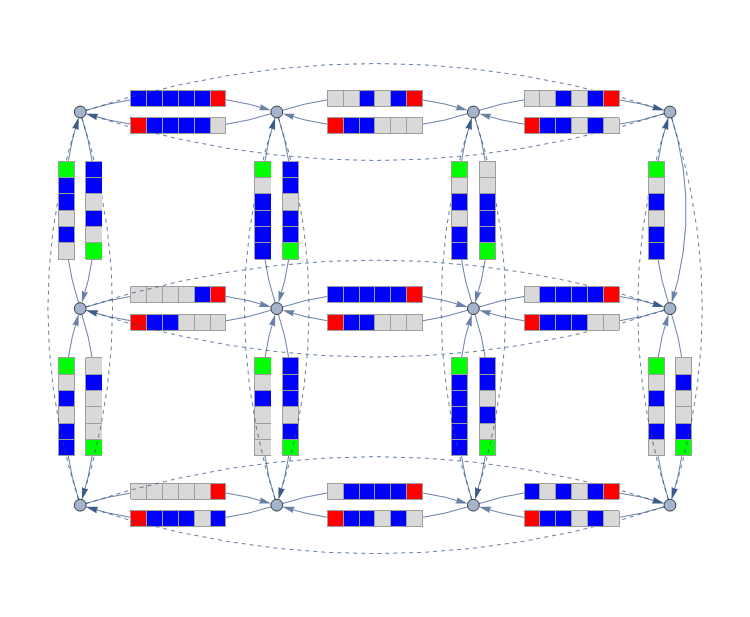}
\caption{Example  $3\times $4 traffic network. The connecting links to form the torus  are shown as dashed directed links; we have omitted the cells on these links to avoid clutter. Each segment has $n=5 $ cells;  an additional   cell has been added downstream of each segment to indicate the traffic light color.}
\label{network}
\end{figure}

\textbf{The signalized network} corresponds to a homogeneous grid network of bidirectional streets, with one lane per direction of length 
\begin{equation}
    n=5\qquad\mbox{cells between neighboring traffic lights.}
\end{equation}
To attain spatial  homogeneity,  the network is defined on a torus:  Street ends  on the edge of the network  are connected  so that each street can be thought of as a ring road. This is illustrated in Fig.~\ref{network},  where we have omitted the cells on the connecting links (to form the torus)  to reduce clutter. Notice that in this setting all intersections have 4 incoming and 4 outgoing approaches.

\textbf{Vehicle routing} is such that reaching the stop line  will choose  to turn right, left  or keep going straight with equal probability. This promotes a uniform distribution for density on the network. 

\textbf{Traffic signals} operate with only two restrictions: a red-red  of one time step (of the CA model) to account for  the lost time steps when switching lights, and a minimum green time, $g$ of 3 time steps. This means that one iteration in the learning framework correspond to  $g$ time steps of the CA model.

\subsection{The DRL framework} 

Each traffic signal is considered an agent that learns from the environment. There are two possible \textbf{actions} for each agent: turning the light red/green for the North-South approaches (and therefore turns the light green/red for the East-West approaches). We don't consider yellow phase in this paper. 
\textbf{The state} observable by the agent is a $8\times n$ matrix of bits, given the four incoming and the four outgoing $c$-vectors from the CA model, one for each approach to the intersection. 

\begin{figure}[tb]
\centering
\includegraphics[width=.7\textwidth]{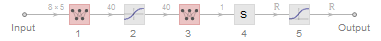}
\caption{Neural network architecture to approximate the policy. The numbers on top of the arrows indicate the dimensions of the corresponding input/output vectors, and the numbers below the squares are as follows: the input  is the state observable by the agent, 1: linear layer, 2: tanh function, 3: linear layer, 4: summation layer, 5: sigmoid function, and
 the output is a single real number that gives the probability of turning the light red for the North-South approaches.}
\label{f3}
\end{figure}

\textbf{The policy}  for each traffic signal agent is approximated by a deep neural network as shown in Fig.~\ref{f3}. It is a 3-layer perceptron with tanh nonlinearity,  known to approximate any continuous function with an arbitrary accuracy provided the network is ''deep enough'' \citep{kuurkova1992kolmogorov}. 
The input to the network is the state observable by the agent, while the output is a single real number that gives the probability of turning the light red for the North-South approaches.

 We define \textbf{the reward} at time $t,\ R_t,$ as the
 \textit{incremental average flow per lane}, defined here as the average flow through the intersection during $(t,t+g)$ \textit{minus} the flow predicted by the network MFD at the prevailing density. 
We will see that this definition of reward is superior to the more standard \textit{average flow per lane} in that case the resulting parameter variance is larger, which makes it more difficult for training algorithms to converge.
In this context the MFD can be seen as a baseline for the learning algorithm, which reduces parameter variance. But for a baseline to be effective it needs to be independent of the actions taken. To this end,  we use the maximum-queue-first (LQF) algorithm as a baseline, whose mean MFD is shown as a thick dashed curve starting in Figure \ref{random}. 

Because our network is spatially homogeneous and without boundaries, there is no reason why policies should be different across agents, and therefore we will train \textit{a single agent} and share its parameters with all other agents. 
After training, we evaluate the performance of the policy  
by observing the resulting MFD. 


\subsection{The training algorithm \ta}

In this paper we propose the training algorithm \ta, which is in the spirit of REINFORCE with baseline \citep{willianms1988toward} but for continuing problems. To the best of our knowledge, this extension of REINFORCE is not available in the literature. Notice that we tried other methods in the literature 
with very similar results, so \ta\  is chosen here since it has the fewest hyper parameters: learning rates $\alpha $ and $\beta$ for the parameters  $\theta $ and the average reward, respectively. Using a grid search over these parameters resulted in $\alpha=0.2 $ and $\beta=0.05$.


Recall that  REINFORCE is probably the simplest policy gradient algorithm that uses \eqref{policy_grad} to guide the parameter search. In the episode setting it is considered a Monte-Carlo method since it requires full episode replay, and it has been considered to be incompatible with continuing problems in literature \citep{sutton2018reinforcement}. 
Here, we argue that a one-step Temporal Difference (TD) approach \citep{sutton1988learning} can be used instead of the Monte-Carlo replay to fit the continuing setting.
This boils down to estimating the differential return \eqref{return} by the temporal one-step differential return  of an action:
\begin{equation}\label{return2}
    G_t\approx R_t-\eta(\pi)
\end{equation}
Notice that the second term in this expression can be interpreted as a baseline in REINFORCE, and baselines are known to reduce parameter variance.
The pseudocode is shown in Algorithm \ref{reinforce}.

\begin{algorithm}
\caption{\ta}
\label{reinforce}
\begin{algorithmic}[1]
\STATE Input: parameterized policy \( \pi (a|s;{\theta }), \theta \in \mathcal{R}^{m} \),  average density $k$
\STATE Set hyper-parameter $\alpha,\beta$, set average reward $\eta=0$
\STATE Initialize vector $\theta$ 
\STATE Initialize the network state $S$ as a Bernoulli process with probability $k$ over the cells in the network 
\REPEAT 
\STATE Generate action $A\sim \pi(\cdot|S;\theta)$ 
\STATE Take action $A$, observe the new state $S'$ and reward $R$ (by running the traffic simulation model for $g$ time steps)
\STATE $G\gets R-\eta$
\STATE $\eta\gets\eta+\beta \ G$
\STATE $\theta\gets\theta+\alpha\  G \ \nabla_{\theta}\log  \pi (A|S;\theta)$
\STATE $S\gets S'$
\UNTIL{forever}
\end{algorithmic}
\end{algorithm}


\section{Experiments} 
In this section we perform a series of experiments to highlight the main properties of our problem. The different policies will be compared based on the MFD they produce  once deployed to all intersections in the network. 
The MFD for each policy is  obtained by  simulating this policy for  several  network densities  and reporting the average flow in the network after  40 time  steps.  This process is repeated  20 times for each density value to obtain an approximate 95\%-confidence interval (mean $\pm$ 2 standard deviations) of the flow for each density value, shown as the shaded areas in all flow-density diagrams that follow.

As a visual benchmark we also use the a greedy method, i.e. longest-queue-first (LQF) policy,  whose mean MFD is shown as a thick dashed curve on the following flow-density diagrams (only the mean values are reported here to avoid clutter). In this way, we are able to test the hypotheses that the policy outperforms LQF simply by observing if the shaded area is above the dashed line.
In particular, we will say that a policy is ``optimal'' if it outperforms  LQF, ``competitive'' if it performs similarly to LQF (shaded area overlaps with dashed line), and ``suboptimal'' if it underperforms LQF (shaded area below the dashed line).

\subsection{Random policies}\label{random section}
\begin{figure}[tb]
\centering
\includegraphics[width=.8\textwidth]{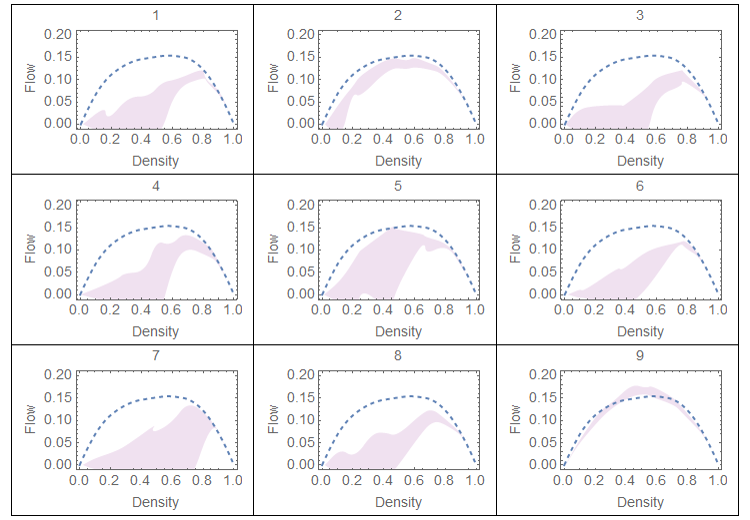}
\caption{Random policies. Each diagram is a different trial, and shows the average density versus average flow in the network. The dashed line corresponds to the benchmark LQF policy. The red and green envelope curves show the MFD bounds. }
\label{random}
\end{figure}

In this experiment the  weights $\theta$ for the policy are set according to a standard normal distribution. As illustrated in Fig.~\ref{random}, it is possible to find a competitive policy after just a few trials;  as in trial 9 in the figure, similar random search method for RL problems can be found in \cite{mania2018simple,camponogara2003distributed}. 
A visual analysis of a large collection of such images reveals that about 15\% of these random policies are competitive. 

Notice that this figure also reveals that all policies, no matter how bad, are optimal when the density exceeds approximately 75\%. To the best of our knowledge, this property of signalized traffic networks has not been reported previously. 
\cite{camponogara2003distributed} was the only study being close to it, which compared average delay by using random search, LQF, and RL-based policies under densities ranging from 0.1 to 0.75. However, they did not show any results under density higher than 0.75, and no explanation or discussion was provided. Although they did not discuss the learning performance under density higher than 0.75, their results given density from 0.1 to 0.75 show consistency with our finding, and supported the density threshold value 0.75 revealed in our study. 

To see the property, consider the upper and lower bounds on the MFD in the figure, as a way of defining a feasibility region for the MFD. The existence of this lower bound is unexpected, and since it overlaps with the upper bound for congested densities means that intersection throughput is not affected by the control.


%
A possible explanation is that under heavily congested conditions there will always be a queue waiting to discharge at all intersections, and therefore which approach gets the green becomes irrelevant. 

\subsection{Supervised learning policies} \label{supervised section}

\begin{figure}[tb]
\centering
\includegraphics[width=\textwidth]{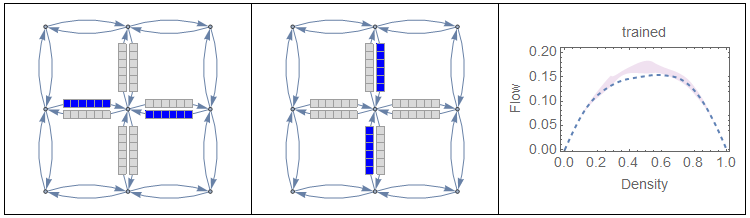}
\caption{Supervised learning experiment. Left: extreme state $s_1$, where both North-South approaches are empty and the East-West ones are at jam density; we have omitted the cells on links other than the ones observable by the middle intersection to avoid clutter. Middle: extreme state $s_2 $, the opposite of $s_1$. Right: resulting MFD (shaded area).}
\label{supervised}
\end{figure}

In this section we will report a rather surprising result, training the policy with \textbf{only two} examples yields a near-optimal policy. These examples are shown in Fig.~\ref{supervised} and correspond to two extreme situations where the choice is trivial: the left panel shows extreme state $s_1$, where both North-South approaches are empty and the East-West ones are at jam density (and therefore red should be given to those approaches with probability one), while the middle panel shows $s_2 $, the opposite situation (and therefore red should be given to North-South approaches with probability zero); in both cases all outgoing approaches are empty. The training data is simply:
\begin{equation}\label{trivial}
    \pi(s_1)\rightarrow 1,\qquad \pi(s_2)\rightarrow 0.
\end{equation}
The figure also shows the MFD resulting from this policy, where it can be seen that it outperforms our benchmark for all densities.

\subsection{DRL policies}


Here we training the policy using DRL, as discussed earlier. We included two experiments. In the first one the policy is trained under a constant number of vehicles in the network, and we show that as soon as congestion builds up the learning process deteriorates. The second experiment considers the standard definition of reward in the literature, and we show that it produces a slower convergence.

\subsubsection{Constant demand} 

In these experiments we consider a constant traffic demand, i.e. the density of vehicles in the network, $k$,  is kept constant during the entire training process. 

\begin{figure}[tb]
\centering
\includegraphics[width=\textwidth]{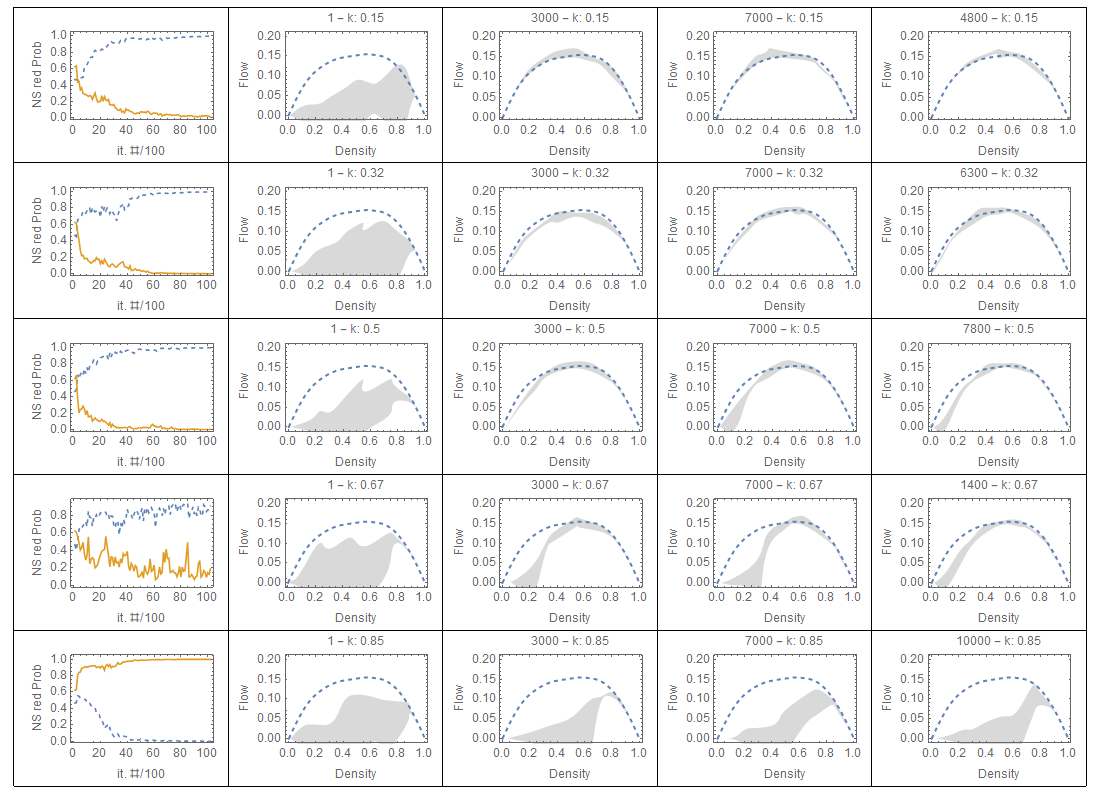}
\caption{Policies trained with constant demand and random initial   parameters $\theta$. The label in each diagram gives the iteration number and the constant density value.  First column: NS red probabilities of the extreme states, $\pi(s_1)$ in dashed line and $\pi(s_2)$ in solid line. The remaining columns show the flow-density diagrams obtained  at different iterations, and the last column shows the  iteration producing the highest flow at $k= 0.5$, if not reported on a earlier column.}
\label{constantr}
\end{figure}

\begin{figure}[tb]
\centering
\includegraphics[width=\textwidth]{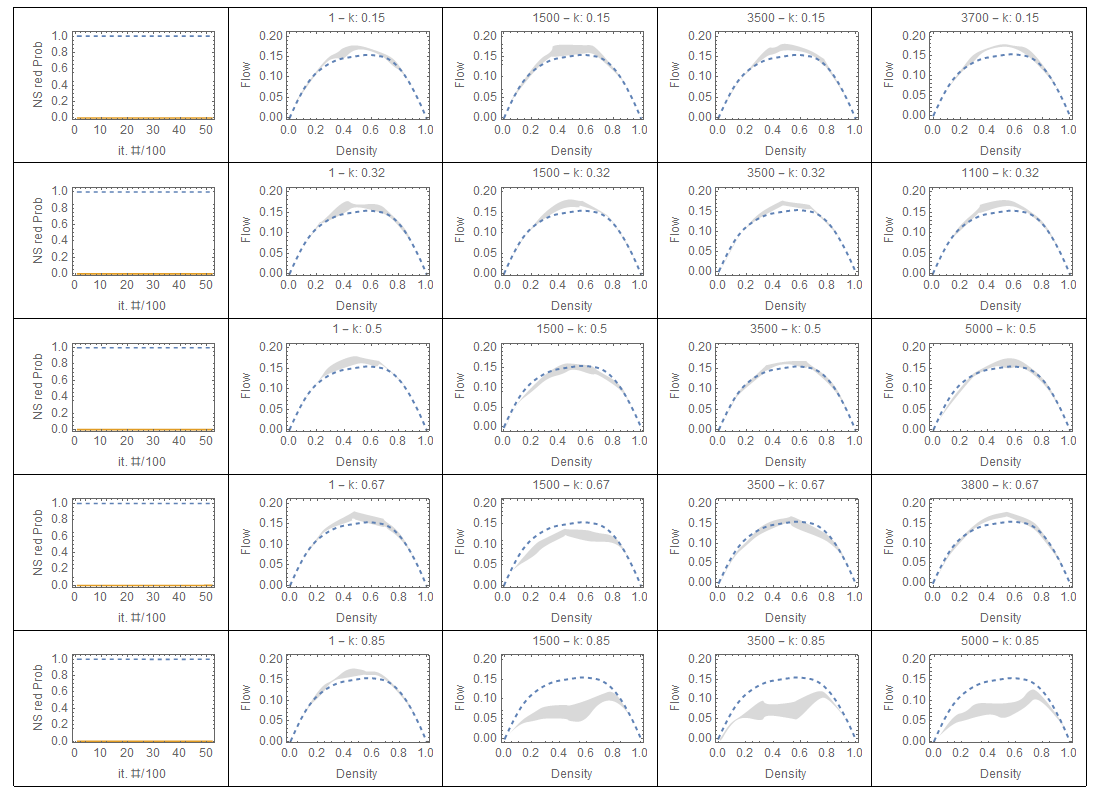}
\caption{Policies trained with constant demand  and supervised initial parameters $\theta$.  The label in each diagram gives the iteration number and the constant density value.  First column: NS red probabilities of the extreme states, $\pi(s_1)$ in dashed line and $\pi(s_2)$ in solid line. The remaining columns show the flow-density diagrams obtained  at different iterations, and the last column shows the  iteration producing the highest flow at $k= 0.5$, if not reported on a earlier column.}
\label{constants}
\end{figure}

The results for three levels of demand, and for random and supervised initial conditions for the policy weights are shown in Fig.~\ref{constantr} and Fig.~\ref{constants}, respectively. Each row corresponds to a constant density level, while the first column depicts the NS red probabilities of the extreme states, $\pi(s_1)$ and $\pi(s_2)$ (described in section \ref{supervised section}) as a function of the iteration number, and these probabilities should tend to \eqref{supervised} for ``sensible'' policies. 
These figures reveals considerable insight:
\begin{enumerate}
    \item the first column in Fig.~\ref{constantr} reveals that a sensible policy cannot be achieved for $k=0.85$. This is apparent because probabilities $\pi(s_1)$ and $\pi(s_2)$ converge to the wrong values. We have verified that for congested traffic conditions with $k\ge0.75$ this result is observed.

    \item for $k\ge 0.5$ in Fig.~\ref{constantr} all the policies obtained are suboptimal and deteriorate as density increases.
    
    \item the best policies in Fig.~\ref{constantr}, albeit only competitive, are obtained for free-flow conditions, i.e. $k\le=0.5$, with lower density leading to slightly better policies.
    \item Fig.~\ref{constants} shows that even starting with initial parameters from the supervised experiment, the additional DRL training under congested conditions, $k\ge 0.5$, leads to a deterioration of the policy.  Under free-flow, conversely, policies seem to improve slightly.

\end{enumerate}

These observations indicate that DRL policies lose their ability to learn and deteriorate as density increases. A possible explanation that deserves further research is elaborated momentarily in the discussion section. We conjecture that this result is a consequence of a property of congested urban networks and has nothing to do with the algorithm to train the DRL policy.

\subsubsection{Non-incremental rewards}

\begin{figure}[tb]
\centering
\includegraphics[width=\textwidth]{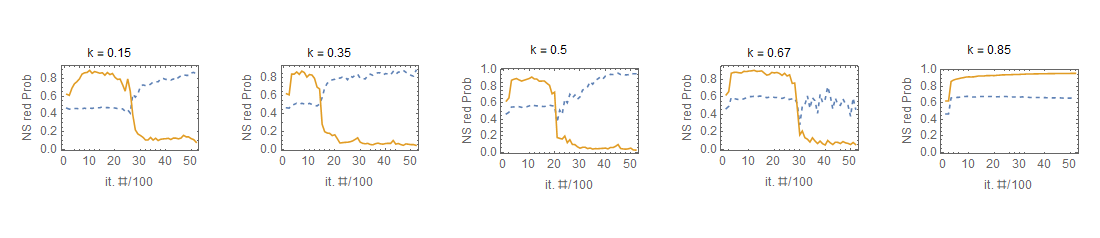}
\caption{Non-incremental rewards with random initial parameters  $\theta$: NS red probabilities of the extreme states, $\pi(s_1)$ in dashed line and $\pi(s_2)$ in solid line.}
\label{non-incremental}
\end{figure}

In this experiment we define {the reward} at time $t,\ R_t,$ as the 
\textit{average flow per lane}  through the intersection during $(t,t+g)$, without subtracting the MFD flow at the prevailing density, as in all previous experiments. This is the standard definition in the literature and we show here that incremental rewards produce faster convergence. This is shown in Fig.~\ref{non-incremental}, which depicts the NS red probabilities $\pi(s_1)$ and $\pi(s_2)$  for random initial parameters  $\theta$,
for the same density levels in the previous experiments.  Comparing these results with the first column in Fig. \ref{constantr} we can see that within the first 2000 or so iterations $\pi(s_1)$ and $\pi(s_2)$ tend to the wrong values but then converge to the expected ones, except for the heavily congested case. 

\section{Discussion and outlook} 

This paper exposed several important properties of machine learning methods applied to traffic signal control on large networks. We have raised more questions than answers at this point, but our future research will focus on formalizing and extending these results.
It is important to note that we have verified that these results remain true for configurations not shown here, i.e. for different (i) number of cells $n$ in each lane and minimum green time $g$, (ii) number of intersections in the network, and (iii) DRL training algorithm.

\begin{figure}[tb]
\centering
\includegraphics[width=0.7 \textwidth]{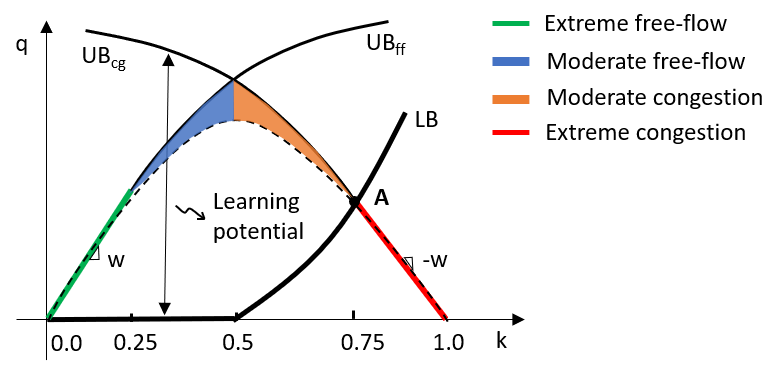}
\caption{Learning potential for DRL in MFD: $UB_{ff}$ denotes the upper bound for free-flow, $UB_{cg}$ denotes the upper bound for congestion, and LB represents the lower bound}
\label{mfd}
\end{figure}

Based on our results and to facilitates this discussion, we argue that networks have 4 distinctive traffic states: extreme free-flow, moderate free-flow, moderate congestion and extreme congestion; see Fig.~\ref{mfd}. To see this, we reason as follows. We know that traffic is symmetric with respect to the critical density: free-flow traffic and congested traffic share the same mathematical properties. In particular, close to the critical density in moderate free-flow and moderate congestion, the network flow exhibits moderate variance, but in extreme free-flow or extreme congestion   the network flow becomes deterministic, as can be confirmed from the numerous simulated MFDs shown here (or from the method of cuts in \citep{Laval2015Stochastic}).

We found a property of congested urban networks, namely \textit{\xx}, that makes DRL methods unable to find sensible policies under congested traffic conditions. We conjecture that this behavior is consistent with the shape of the MFD bounds uncovered in section \ref{random section}, whereby the more the congestion, the less the policy affects   intersection  throughput. This tendency of the flow to be independent of the policy under congestion renders gradient information less meaningful, which corrupts the learning process.
Even starting with initial weights given by the supervised training policy, we saw that additional training under congested conditions leads to a deterioration of the policy. Similarly, we have verified similar behavior  under dynamic demands whenever congestion appears in the network.

To the best of our knowledge, this behavior has been mentioned only once in the literature \cite{de2006reinforcement}, but no follow-up research has been generated since. This is unfortunate because this means, potentially, that all the DRL methods proposed in the literature to date are unable to learn as soon as congestion appears on the network. It also means that the limited success of DRL for traffic signal control might be explained by \xx, which has been overlooked so far. But the current explanation in the literature to explain the limited success of DRL is that the problem is non-stationary and/or non-Markovian \cite{choi2000hidden,da2006dealing}, which probably explain why we still have not solved the problem.

It is important to investigate new DRL methods able to cope with \xx, i.e. methods able to extract relevant knowledge from congested conditions. 
In the meantime, it is advisable to train DRL policies under free-flow conditions only, discarding any information from heavily congested ones. We have shown here that such policies are nearly optimal for all traffic conditions.
This intriguing result indicates that most of what the agent needs to learn is encoded in free-flow conditions, and that the data in congestion is irrelevant. We will challenge this idea in future studies with networks that are not ``ideal'' as in here. 

\begin{figure}[tb]
\centering
\includegraphics[width=.495\textwidth]{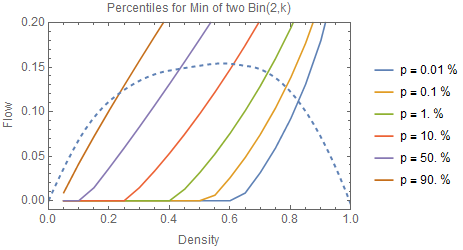}
\includegraphics[width=.495\textwidth]{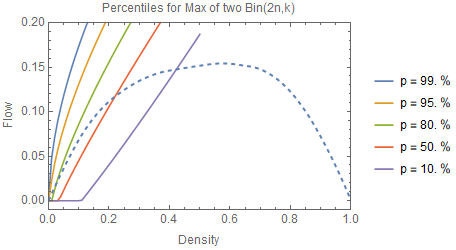}
\caption{Percentiles for the minimum (left) and maximum (right) of two independent binomial random variables with parameters $2n$ and $k$.}
\label{quantic}
\end{figure}

We suspect, that \xx will still hold for general networks because every network should have a lower bound for the MFD that is greater than zero under congestion. To see this, consider a very bad policy that gives the green to the smallest queue at every time instant. In free-flow conditions it is likely that the smallest queue will be an empty queue and therefore the throughput would be zero. But under congestion the smallest queue will tend to be greater than zero and therefore the throughput has no choice but to increase. 
We can formalize this idea by assuming that at each time instant the number of vehicles in the network is described by a Bernoulli process of probability $k$. It follows that the number of vehicles in the NS approaches is binomial with parameters $n_1+n_2$ and $k$, where $n_1, n_2$ are the number of cells of both NS approaches, respectively. And similarly for the distribution of the number of vehicles in the  EW approaches. 
This can be used to obtain the flow through the intersection (provided outgoing approaches do not block traffic) simply by dividing by $2n$ to obtain the density and multiplying by the free-flow speed of 1 to obtain the flow.
Therefore, the distribution of the number of vehicles flowing through the intersection under this bad policy is simply linear transformation of the distribution of the minimum of these two binomial random variables.
It turns out that the percentile function of this distribution for small probabilities exhibits the desired shape for the MFD lower bound. The left panel in
Fig.~\ref{quantic} (left) shows this function for $n_1=n_2=n$ and for selected probabilities along with the MFD for the LQF policies for reference, where the infeasible region for the MFD becomes apparent. Similar behavior is observed for different values of $n_1,n_2$, not shown here for brevity, and therefore these lower bounds should exist in general networks.


For the upper bounds, the method of cuts provides the answer, and the reader is referred to \citep{daganzo2008analytical, Laval2015Stochastic} for the details. 
Given this upper bound, it becomes clear that the congested states beyond the intersection point ``A'' in  Fig.~\ref{mfd} obey very different rules compared to congestion near capacity. In this extreme congestion state the control has absolutely no influence on network flow, which is deterministic in which explains the failure of DRL methods to learn (because there is nothing to learn!). In moderate congestion closer to the critical density the flow is stochastic and the lower bound starts activating, causing the distance between upper and lower bounds to decrease with density, which might explain the learning difficulties in this traffic state.  Moderate free-flow is similar, except that the lower bound remains at zero flow. Finally, extreme free-flow is different than extreme congestion because in the former the lower bound is zero.

The above paragraph suggests that at any given density the distance between the upper bound in congestion and the lower bound is a measure of the potential for DRL methods to learn. This ``learning potential'' is maximum in free-flow and starts shrinking 
until disappearing at point ``A'' in  Fig.~\ref{mfd}. But in practice this learning potential might be much narrower than shown in the figure, which explains why learning difficulties start near the critical density. To see this, consider the median outflow in Fig.~\ref{quantic} (left), which  indicates that most of the time a bad policy will produce remarkably high flows; e.g. for $k=0.5$ the median outflow is around 0.2.
But this high outflow will be above the upper bounds in Fig.~\ref{mfd}. This means that most of the time the throughput of an intersection is dictated by the infrastructure (the upper bounds) rather than the policy.
This is the simplest explanation consistent with the results in this paper, but research is needed to understand how to use it for learning under congestive conditions.

Notably, we also found that supervised learning with only two examples yields a near-optimal policy. This intriguing result indicates that  extreme states $s_1$ and $s_2$ encode vital information and that the neural network can successfully extrapolate to all other states. (Notice however, that this result cannot be obtained if the input to the neural network is a $4\times n$  matrix instead of the $8\times n$ matrix used in this paper.) Understanding precisely why this happens could lead to very effective supervised learning methods based on expert knowledge, and to supplement DRL's inability to learn under congested conditions.


Combining the results in this paper with those in \cite{Laval2015Stochastic} we conjecture that a necessary condition for a policy to be optimal under congestion is that the average green time given to any incoming approach be proportional to the length of the approach.  To see this, recall that  \cite{Laval2015Stochastic} show that the MFD can be well approximated by a function of mainly two  parameters: $\lambda=$ the mean distance between traffic lights divided by the mean green time, and $\rho= $ the mean red-to-green ratio across the network.  Since our network is spatially homogeneous we have $\rho= 1$. Using this  in equation (17b) of \cite{Laval2015Stochastic} we infer that in the deterministic case ($\delta= 0$ in (17b)) the slope of the MFD in extreme congestion, namely $-w$; see Fig.~\ref{mfd}, is given by:
\begin{equation}
    w=\frac{\lambda}{1/2+\lambda}.
\end{equation}
The reader can verify from the many MFD's shown  here that $w\approx2/3$ and therefore that $\lambda\approx 1$. This means that the mean green time produced by the policy matches the mean distance between traffic lights (in dimensionless form)busy working so. Notice that $\lambda< 1$ was shown in \cite{Laval2015Stochastic} to be the short-block condition, i.e. the network becomes prone to spill back, which can have a severe effect on capacity. Conversely, a network with $\lambda>1$ has  long blocks (compared to the green time) and therefore will not exhibit spill back. Therefore, that an optimal policy produce  $\lambda< 1$ is not surprising as it indicates that green times are just long enough  as to not produce spill back.
This highlights the importance of considering segment length when deciding signal timing, a subject rarely mentioned in the signal control literature.





\section{Acknowledgements}
This study has received funding from NSF research projects \# 1562536 and \# 1826162.

\bibliographystyle{trb.bst}
\bibliography{bibnew}

\nolinenumbers
\end{document}

%% file: pream.tex
\usepackage[tiny,rm]{titlesec}

\newpagestyle{trbstyle}{
\sethead{Laval, and Zhou}{}{\thepage}
}
\titleformat{\section}{\bfseries}{}{0pt}{\uppercase}
\titlespacing*{\section}{0pt}{12pt}{*0}
\titleformat{\subsection}{\bfseries}{}{0pt}{}
\titlespacing*{\subsection}{0pt}{12pt}{*0}
\titleformat{\subsubsection}{\itshape}{}{0pt}{}
\titlespacing*{\subsubsection}{0pt}{12pt}{*0}

\usepackage[inline]{enumitem}
\setlist[1]{labelindent=0.5in,leftmargin=*}
\setlist[2]{labelindent=0in,leftmargin=*}

\usepackage{ccaption}
\usepackage{amsmath}
\usepackage{empheq}
\usepackage{multirow}
\usepackage{amssymb}
\makeatletter
\renewcommand{\fnum@figure}{\textbf{FIGURE~\thefigure} }
\renewcommand{\fnum@table}{\textbf{TABLE~\thetable} }
\makeatother
\captiontitlefont{\bfseries \boldmath}
\captiondelim{\;}

\usepackage{mathptmx}

\usepackage[T1]{fontenc}
\usepackage{textcomp}

\usepackage[sort,numbers]{natbib}

\setcitestyle{round}

\usepackage[pagewise]{lineno}

\newread\somefile
\usepackage{xparse}

\newcounter{totalwordcounter}
\newcounter{wordcounter}
\makeatletter

\NewDocumentCommand{\wordcount}{s}{%
  \immediate\write18{texcount -sum -1 \jobname.tex > count.txt}%
  \immediate\openin\somefile=count.txt%
  \read\somefile to \@@localdummy%
  \immediate\closein\somefile%
  \setcounter{wordcounter}{\@@localdummy}%
  \IfBooleanF{#1}{%
  \@@localdummy
  }%
}
\makeatother
\usepackage{totcount}
	\regtotcounter{table} 	
	\regtotcounter{figure} 	

\newcommand{\wordtable}{250} 

\newcommand{\totalwordcount}{%
  \wordcount*
  \setcounter{totalwordcounter}{\value{wordcounter}}%
  \addtocounter{totalwordcounter}{\numexpr\wordtable*\totvalue{table}} %
  \number\value{totalwordcounter}
  \renewcommand{\totalwordcount}{\number\value{totalwordcounter}}
}

\usepackage{graphicx}
